\documentclass[journal]{IEEEtranTIE}
\usepackage{graphicx}
\usepackage{cite}
\usepackage{picinpar}
\usepackage{amsmath}
\usepackage{url}
\usepackage{flushend}
\usepackage[latin1]{inputenc}
\usepackage{colortbl}
\usepackage{soul}
\usepackage{multirow}
\usepackage{pifont}
\usepackage{color}
\usepackage{alltt}
\usepackage[hidelinks]{hyperref}
\usepackage{enumerate}
\usepackage{siunitx}
\usepackage{breakurl}
\usepackage{epstopdf}
\usepackage{pbox}

\usepackage{epsfig} %
\usepackage{mathptmx} %
\usepackage{eucal}
\usepackage{times} %
\usepackage{amssymb}  %
\usepackage{graphicx}
\usepackage{soul}
\usepackage{xcolor}
\sethlcolor{yellow}
\usepackage{dsfont}
\usepackage{xcolor} %
\usepackage{booktabs}
\usepackage{makecell}
\usepackage{multirow}
\usepackage{balance}
\usepackage{fancyhdr}
\fancypagestyle{accepted}{
    \fancyhf{}
    \fancyhead[C]{%
        \parbox{\textwidth}{%
            \centering
            \footnotesize
            \textbf{This is the accepted version of an article in
            IEEE Transactions on Industrial Electronics, 2026.\\}
            Final published version (DOI): 10.1109/TIE.2026.3748262.\\
            Multimedia: \url{https://youtu.be/Y7SBvGf0Ah8}
        }%
    }
    \fancyfoot[C]{%
        \parbox{\textwidth}{%
            \centering
            \fontsize{6}{7}\selectfont
            \copyright\ 2026 IEEE. Personal use of this material is permitted.
            Permission from IEEE must be obtained for all other uses,
            in any current or future media, including
            reprinting/republishing this material for advertising
            or promotional purposes, creating new collective works,
            for resale or redistribution to servers or lists, or
            reuse of any copyrighted component of this work in other works.
        }%
    }

}

\begin{document}
\pagestyle{accepted}
\title{Reinforcement Learning Based Multi-Terrain Adaptive Whole-Body Control for a Quadruped Manipulator}

\author{
	Yanzhou Jin$^{1}$, Fei Suo$^{1}$, Chang Liu$^{1}$, Yasuhisa Hasegawa$^2$, \emph{Member, IEEE}, Liming Chen$^3$, \emph{Senior Member, IEEE}, Yusuke Iwasawa$^1$, Yutaka Matsuo$^1$ and Yaonan Zhu$^{1,*}$, \emph{Member, IEEE}
    \thanks{This work was supported in part by JST PRESTO, Japan, under Grant JPMJPR25T5; JST SICORP, Japan, under Grant JPMJSC24C5; JSPS KAKENHI under Grant JP24K17236; and the Nagamori Foundation. (*Corresponding author: Yaonan Zhu)}
	\thanks{$^{1}$Graduate School of Engineering, The University of Tokyo, Japan. (email: \{yanzhou.jin, fei.suo, chang.liu, iwasawa, matsuo, yaonan.zhu\}@weblab.t.u-tokyo.ac.jp) 

$^{2}$Department of Micro-Nano Mechanical Science and Engineering, Nagoya University, Nagoya, 464-8603, Japan. (email: hasegawa@mein.nagoya-u.ac.jp)

$^{3}$Department of Mathematics and Informatics, Ecole Centrale de Lyon, France. (email: liming.chen@ec-lyon.fr)}

}

\maketitle
\thispagestyle{accepted}
\begin{abstract}
Legged robots equipped with manipulators offer promising capabilities for loco-manipulation tasks in unstructured environments. However, achieving task-consistent end-effector behavior during locomotion remains challenging due to terrain-induced disturbances and the strong coupling between locomotion and manipulation, particularly for small-scale quadruped platforms. A reinforcement-learning-based whole-body control framework is proposed for quadruped loco-manipulation on complex terrain. 
The proposed framework combines a hybrid task representation with task-prioritized whole-body learning to coordinate locomotion and manipulation while preserving task-consistent end-effector behavior. A lightweight quadruped manipulator platform is further developed to validate the proposed framework on a small-scale quadruped system.
Real-world experiments demonstrate successful deployment on complex terrains such as slopes up to 30 degrees, stairs, and deformable terrain. These results demonstrate the feasibility of deploying a learned whole-body control policy for quadruped loco-manipulation in real-world environments and highlight the importance of task-consistent whole-body coordination for terrain-adaptive loco-manipulation. 

\end{abstract}


\definecolor{limegreen}{rgb}{0.2, 0.8, 0.2}
\definecolor{forestgreen}{rgb}{0.13, 0.55, 0.13}
\definecolor{greenhtml}{rgb}{0.0, 0.5, 0.0}

\section{Introduction}
Recent advances in quadruped locomotion have enabled robust traversal across challenging terrains. By integrating onboard manipulators, quadruped robots are expected to perform loco-manipulation tasks that require both mobility and object interaction, such as inspection, maintenance, search-and-rescue, and operation in unstructured environments. However, combining locomotion and manipulation introduces strong coupling between base motion and end-effector (EE) behavior, making precise task execution substantially more challenging than pure locomotion \cite{An2025Collaborative, ma2025learning, rahman2026loco}.

Compared with larger mobile manipulators, small-scale quadruped manipulators operate under significantly tighter payload constraints \cite{ferrolho2023roloma,dadiotis2023whole,sambhus2026nonlinear}. Commercially available arms used on such platforms can occupy a substantial portion of the quadruped payload capacity and act as moving loads during locomotion \cite{wang2025odysseyopenworldquadrupedsexploration,ha2024umi}. Their static mass, configuration-dependent center-of-mass shifts, and dynamic reactions may increase leg actuator demands and reduce the performance of terrain traversal.
Conversely, terrain irregularities and changing foothold contacts can induce body motions that are transmitted to the end effector, thereby compromising manipulation performance \cite{rahman2026loco}. 
Existing loco-manipulation systems often adjust body posture to expand the manipulator workspace and improve end-effector reachability \cite{portela2024learning,xie2024manipulability}. Although effective on flat terrain, such posture adjustments may conflict with the body regulation required for terrain traversal \cite{lee2021whole}, whereas body motions required for terrain adaptation may disturb end-effector tracking \cite{zhou2022teleman,zhang2023dynamic}. Accordingly, maintaining task-relevant end-effector behavior while preserving locomotion feasibility on challenging terrain remains a central challenge for quadruped loco-manipulation.

Reliable locomotion requires adequate body regulation \cite{lee2021whole}, while manipulation depends on maintaining locomotion feasibility. These dependencies motivate the state-dependent prioritization of body regulation, locomotion, and manipulation. Existing loco-manipulation approaches commonly coordinate locomotion and manipulation through unified optimization frameworks \cite{molnar2025whole,risiglione2025rakomo,sleiman2021unified}, hierarchical locomotion-manipulation architectures \cite{turrisi26littercollection}, or learning-based whole-body control (WBC) policies \cite{hou2025efficient,jiang2024learning}. Optimization-based approaches can explicitly incorporate whole-body dynamics and physical constraints \cite{bellicoso2017dynamic}, but their application to rough terrain depends on sufficiently accurate modeling of coupled robot dynamics, changing contacts, and terrain interactions, together with repeated online optimization. Reinforcement-learning (RL) is well suited to this setting because it allows closed-loop whole-body coordination to be optimized offline over randomized terrain, contact, and loading conditions, while deployment requires only fixed-cost policy inference. Recent learning-based studies have demonstrated whole-body trajectory tracking \cite{ha2024umi}, object interaction \cite{liu2024visual,liu2025mlm}, and mobile manipulation \cite{fu2023deep,zhang2025learning,liang2026whole}. However, explicit state-dependent coordination among body regulation, locomotion, and manipulation has received limited attention, particularly regarding how manipulation should be modulated when body regulation or locomotion deteriorates. Incorporating these objective dependencies into a unified learned policy therefore remains an open challenge for rough-terrain loco-manipulation.

Beyond objective coordination, end-effector position and orientation play different roles in manipulation tasks. Positional deviations can often be compensated through robot repositioning or locomotion planning, whereas orientation tracking frequently has a direct impact on task execution. Examples include container transportation, tool manipulation, and other disturbance-sensitive tasks in which the end-effector orientation must remain consistent relative  to the environment \cite{Heins_2023}. Consequently, treating end-effector position and orientation identically may lead to unnecessary compromises between task consistency and locomotion feasibility. Effective task representations that balance these competing requirements, therefore, remain another key challenge for rough-terrain loco-manipulation.

\begin{figure}
    \centering
    \includegraphics[width=0.95\linewidth]{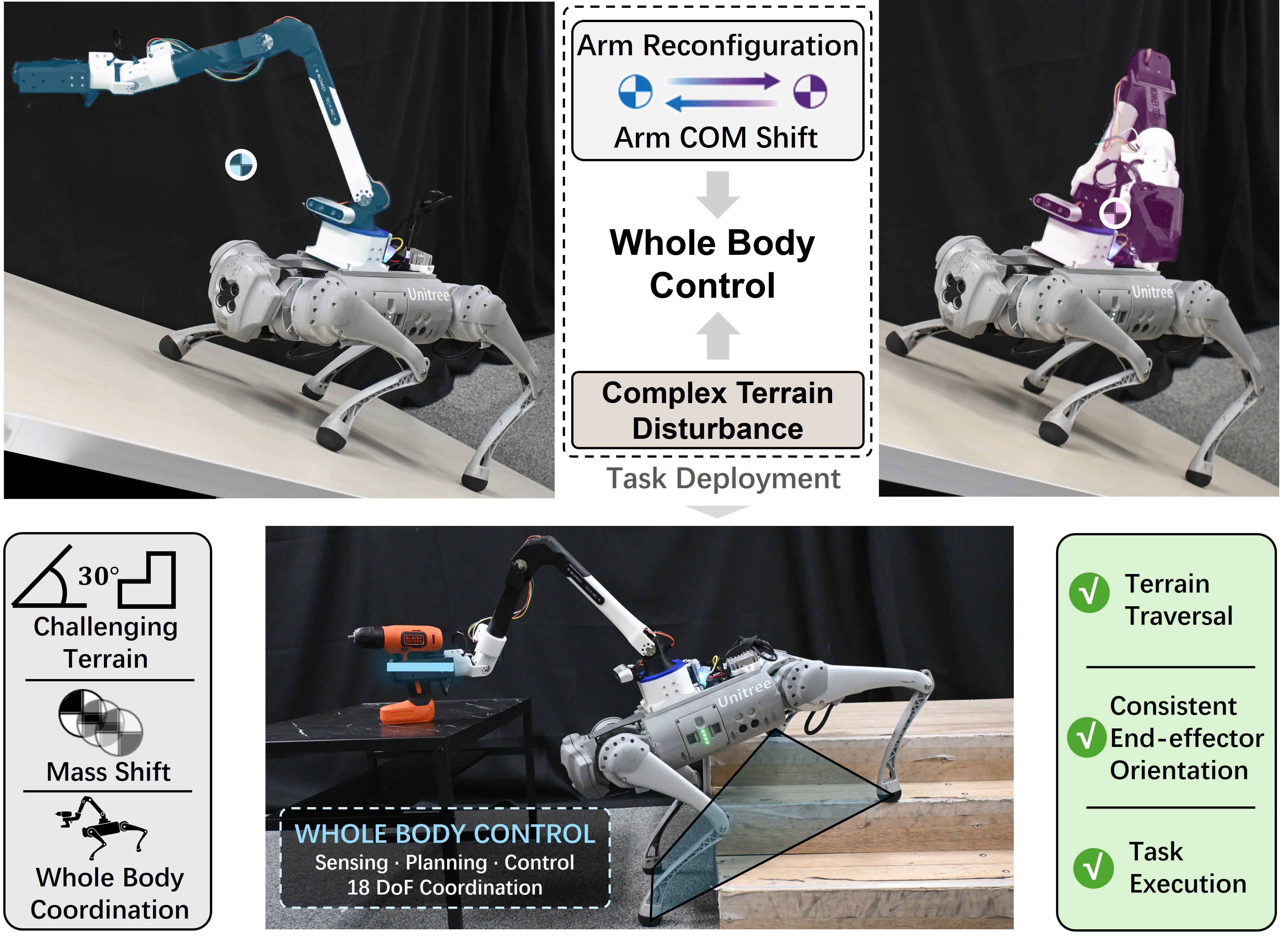}
    \caption{Manipulator-induced mass shifts and complex terrain interaction jointly challenge rough-terrain loco-manipulation on small-scale quadruped robots. The proposed whole-body control framework enables coordinated terrain traversal and manipulation while preserving task-consistent end-effector behavior.}
    \label{fig:teaser}
    \vspace{-4mm}
\end{figure}

To address these challenges, a reinforcement-learning-based whole-body loco-manipulation framework combining task-prioritized learning with a hybrid task-space representation is proposed. The policy is optimized using  Proximal Policy Optimization (PPO~\cite{schulman2017proximalpolicyoptimizationalgorithms}) because its clipped update objective supports stable training, while its rollout-based formulation is compatible with large-scale parallel simulation for continuous 18-DoF control under randomized terrain and contact conditions. Task-prioritized learning modulates the contributions of lower-priority objectives according to the current performance of body regulation and locomotion. The hybrid representation regulates end-effector orientation relative to gravity to preserve task consistency, while expressing end-effector position in the body frame to maintain tracking feasibility under terrain-induced body motions. A lightweight quadruped manipulator is further constructed by integrating a custom-designed 1.5-kg Lite Arm with a Unitree Go1 (Fig. \ref{fig:teaser}). The lightweight design reduces manipulator-induced loading, while the whole-body policy coordinates the arm and base under the remaining configuration- and motion-dependent coupling. Extensive simulation and real-world experiments demonstrate consistent end-effector tracking on challenging terrains, including slopes, stairs, and deformable surfaces.
The contributions of this work are summarized as follows:

\begin{itemize}

\item A lightweight quadruped manipulator platform integrating a custom-designed Lite Arm and a Unitree Go1 is developed, together with a hybrid task-space representation that enables rough-terrain loco-manipulation while preserving task-consistent end-effector behavior.

\item A reinforcement-learning-based whole-body control framework with task-prioritized learning is proposed, enabling coordinated loco-manipulation while maintaining end-effector task tracking under challenging terrain. A systematic ablation study validates the effectiveness of the proposed objective hierarchy.

\item The proposed whole-body control framework is validated through simulation and real-world experiments on challenging terrains, including slopes, stairs, and deformable surfaces. The controller is further integrated into an autonomous loco-manipulation pipeline, demonstrating practical deployment in real-world tasks.

\end{itemize}

\section{Method}

\subsection{System Configuration and Problem Formulation}

\begin{figure*}
    \centering
    \includegraphics[width=\linewidth]{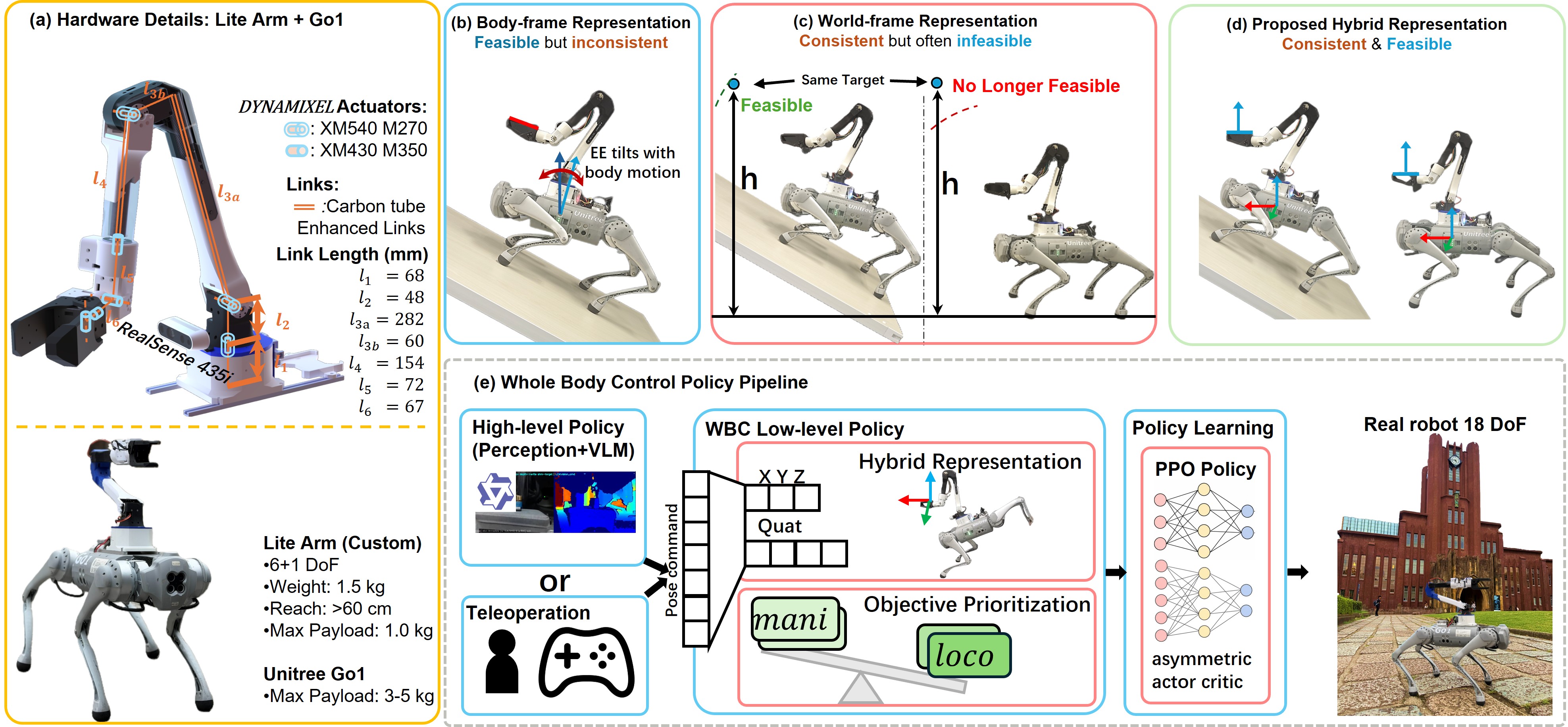}
    \caption{Overview of the proposed framework. (a) Customized lightweight 6-DoF manipulator and the quadruped manipulator system for challenging terrain loco-manipulation. (b)-(d) Comparison of body-frame, world-frame, and hybrid task representations for rough-terrain loco-manipulation. (e) System architecture comprising semantic perception, hybrid task representation, prioritized whole-body learning, and real-world deployment.}
    \vspace{-3mm}
    \label{fig:quadm}
\end{figure*}

The key problem is to preserve both locomotion feasibility and task-consistent end-effector behavior during rough-terrain traversal. To reduce manipulator-induced interference with locomotion at the hardware level, a lightweight 6-DoF manipulator was developed, as shown in Fig.~\ref{fig:quadm} (a). The manipulator weighs only 1.5 kg and provides a maximum reach exceeding 60 cm. Its structure combines carbon-fiber-reinforced polylatic acid (PLA-CF) components with embedded carbon-fiber tubes to provide sufficient stiffness without imposing excessive additional mass. Based on this platform, rough-terrain loco-manipulation is formulated as a unified whole-body control problem involving body regulation, locomotion tracking, and end-effector regulation.

The control problem is formulated as a command-conditioned Markov decision process (MDP), where the policy follows a whole-body command $\mathbf{c}_t^{\mathrm{WBC}}$. A whole-body policy $\pi_\theta$ is optimized by PPO to maximize the expected cumulative objective $\mathcal{J}(\theta)$:

\begin{equation}
\underset{\theta}{\operatorname{maximize}}
\quad
\mathcal{J}(\theta)
=
\mathbb{E}_{\pi_\theta}
\left[
\sum_{t=0}^{T-1}
r_t^{\mathrm{pri}}
\right].
\label{eq:learning_formulation}
\end{equation}

The whole-body command $\mathbf{c}_t^{\mathrm{WBC}}$ and the task-prioritized reward $r_t^{\mathrm{pri}}$ address objective conflicts at different levels. 
At the task-definition level, $\mathbf{c}_t^{\mathrm{WBC}}$ combines the locomotion command with a hybrid EE task representation that balances local reachability and task consistency. 
At the learning level, $r_t^{\mathrm{pri}}$ provides a stepwise learning signal that scales the contributions of locomotion and manipulation based on the current performance of higher-priority objectives.
The resulting reward sequence is used to compute the returns and advantages for the PPO policy update.

\subsection{Hybrid Manipulation Task Representation}
Many manipulation tasks require the end-effector (EE) to maintain a task-consistent pose with respect to the environment. 
For example, when carrying containers or fragile objects, the EE orientation should remain approximately aligned with gravity, which favors a world-frame representation. However, defining the entire manipulation task in the world frame requires reliable global pose estimation and may compromise kinematic feasibility as body motion changes the manipulator workspace relative to the target. This issue is particularly significant for lightweight 6-DoF manipulators with limited workspace and without kinematic redundancy.

Conversely, defining the entire manipulation task in the body frame improves kinematic feasibility but sacrifices task consistency. For example, when the robot traverses a 30$^\circ$ slope, an EE orientation defined purely in the body frame would tilt together with the robot body, which is undesirable for many practical manipulation tasks (Fig.~\ref{fig:quadm}(b)--(d)).

To address this trade-off, the complete command provided to the WBC policy comprises a locomotion command and a hybrid EE command:
\begin{equation}
\mathbf{c}_t^{\mathrm{WBC}}
=
\left(
\mathbf{c}_t^{\mathrm{loco}},
\mathbf{c}_t^{\mathrm{EE}}
\right),\quad
\mathbf{c}_t^{\mathrm{EE}}
=
\left(
{}^{B}\mathbf{p}_{t},
\mathbf{q}_{t}
\right).
\end{equation}
Here, $\mathbf{c}_t^{\mathrm{loco}}$ specifies the desired linear velocities along the $x$- and $y$-axes and the desired yaw angular velocity, while $\mathbf{c}_t^{\mathrm{EE}}$ specifies the hybrid EE command. 
Specifically, ${}^{B}\mathbf{p}_{t}$ denotes the EE position command expressed in the robot body frame, whereas $\mathbf{q}_{t}$  denotes the desired EE orientation, represented as a unit quaternion, such that the EE z-axis is aligned with the world vertical direction obtained from gravity.
Expressing the position target in the body frame allows the robot body to perform large-scale locomotion while the manipulator focuses on local task execution. It also helps maintain local reachability during terrain traversal without requiring global position estimation. Meanwhile, the gravity-referenced orientation prevents the EE from tilting together with the robot body.

Although the hybrid EE representation reduces the conflict between local reachability and task consistency at the task-definition level, multiple whole-body objectives can still impose conflicting demands during policy learning. The following subsection addresses these remaining conflicts through task-prioritized whole-body learning.

\subsection{Task-Prioritized Whole-Body Learning} \label{prioritize}

The hybrid manipulation task representation specifies the desired end-effector behavior, but body regulation, locomotion, and manipulation may still produce competing learning objectives. Locomotion feasibility depends on adequate body regulation, while manipulation during terrain traversal depends on maintaining sufficient locomotion performance. Based on these dependencies, the whole-body objectives are organized according to the following hierarchy:
\vspace{-1mm}
\begin{equation}
\text{Body Regulation}
\succ
\text{Locomotion}
\succ
\text{Manipulation}
\end{equation}

The hierarchy is implemented through performance-conditioned activation factors that scale lower-priority objective-group rewards at each control step. For two objectives $A\succ B$, the current performance of $A$ determines the activation factor applied to $B$. Thus, $B$ is attenuated when the performance of $A$ is insufficient and becomes fully active once the required level is reached.

\subsubsection{Priority I: Body Regulation}

The body-regulation objective $R_{\mathrm{body}}$ has the highest priority because excessive roll or pitch can compromise terrain traversability and the execution of the two lower-priority objectives. Its current condition determines the activation factor $\lambda_{\mathrm{upright}}$, which regulates both locomotion and manipulation:
\begin{equation}
\lambda_{\mathrm{upright}}
=
\min\left(
\frac{\max\left(-\mathbf{g}^{\top}\mathbf{e}_z,\,0\right)}{c_0},
\,1
\right)
\end{equation}
where $\mathbf{g}$ denotes the projected gravity vector expressed in the robot body frame, $\mathbf{e}_z=[0,0,1]^\top$ denotes the body-frame vertical direction, and $c_0>0$ is the upright-alignment threshold applied to the nonnegative score $\max(-\mathbf{g}^{\top}\mathbf{e}_z,0)$. Rather than an additional task objective, $\lambda_{\mathrm{upright}}\in[0,1]$ regulates the contributions of locomotion and manipulation at each control step. As the body deviates from the desired upright configuration, both contributions are reduced accordingly.

\subsubsection{Priority II: Locomotion Regulation}

The locomotion objective $R_{\mathrm{loco}}$ has the second priority because manipulation during terrain traversal relies on sufficient locomotion performance. Its current value determines the activation factor $\lambda_{\mathrm{loco}}$, which further regulates manipulation:
\begin{equation}
\lambda_{\mathrm{loco}}
=
\min\left(
R_{\mathrm{loco}}/c_1,
\,1
\right)
\end{equation}

where $R_{\mathrm{loco}}$ denotes the current nonnegative locomotion-performance objective, and $c_1>0$ is the locomotion-performance threshold at which $\lambda_{\mathrm{loco}}$ saturates at one. Accordingly, $\lambda_{\mathrm{loco}}\in[0,1]$ reduces the influence of manipulation when locomotion performance is insufficient, while retaining its full contribution when the required locomotion capability is maintained.

\subsubsection{Priority III: End-Effector Behavior Regulation}

The manipulation objective $R_{\mathrm{mani}}$ regulates the end-effector behavior specified by the hybrid task representation and is defined as
$R_{\mathrm{mani}}
=
R_{\mathrm{pos}}^{\mathrm{coarse}}
+
R_{\mathrm{pos}}^{\mathrm{fine}}
+
R_{\mathrm{ori}}$,
where $R_{\mathrm{pos}}^{\mathrm{coarse}}$ provides a coarse target-reaching objective based on the Euclidean distance between the commanded and measured end-effector positions, denoted by $e_{\mathrm{pos}}$. To further improve local tracking accuracy, $\mathbf{R}_{\mathrm{pos}}^{\mathrm{fine}}$ is defined as
\begin{equation}
R_{\mathrm{pos}}^{\mathrm{fine}}
=
1-\tanh
\left(
\frac{e_{\mathrm{pos}}}{\sigma}
\right),
\qquad
\left|
\frac{\partial R_{\mathrm{pos}}^{\mathrm{fine}}}
{\partial e_{\mathrm{pos}}}
\right|
=
\frac{1}{\sigma}
\operatorname{sech}^{2}
\left(
\frac{e_{\mathrm{pos}}}{\sigma}
\right),
\end{equation}

where $\sigma>0$ determines the effective range of the fine objective. Its sensitivity to $e_{\mathrm{pos}}$ decreases rapidly as the error increases, concentrating its effect near the target. Thus, $R_{\mathrm{pos}}^{\mathrm{fine}}$ provides the fine-position regulation within Priority III.

\subsubsection{Prioritized Objective Integration}

The three objective groups are integrated through the state-dependent priority vector
\begin{equation}
\boldsymbol{\Lambda}
=
\begin{bmatrix}
1,~
\lambda_{\mathrm{upright}},~
\lambda_{\mathrm{upright}}\lambda_{\mathrm{loco}}
\end{bmatrix}^{\top}
\end{equation}
whose elements correspond to body regulation, locomotion, and manipulation, respectively. The stepwise prioritized objective is used as the reward signal and is defined as
\begin{equation}
r_t^{\mathrm{pri}}
=
\boldsymbol{\Lambda}^{\top}
\textbf{R},
\quad
\textbf{R}
=
\left[
R_{\mathrm{body}},\;
R_{\mathrm{loco}},\;
R_{\mathrm{mani}}
\right]^{\top},
\end{equation}
where each objective group $R_{(\cdot)}$ is constructed from the corresponding task-specific reward components described in Section~\ref{reward}. 
Since $\boldsymbol{\Lambda}$ is updated from the current body condition and locomotion performance, it schedules the relative contributions of the objective groups throughout each rollout. The resulting sequence of $r_t^{\mathrm{pri}}$ is used to compute the returns and advantages through which PPO updates a single shared policy.

By maximizing the expected return defined by $r_t^{\mathrm{pri}}$, PPO encourages the shared policy to coordinate the three objective groups under the prescribed hierarchy. The resulting priority is soft rather than a strict non-interference constraint. When higher-priority performance falls below the required level, the corresponding activation factor decreases, reducing the influence of dependent lower-priority objectives in $r_t^{\mathrm{pri}}$.

The integration of the hybrid task representation and objective prioritization within the overall WBC framework is summarized in Fig.~\ref{fig:quadm}(e).

\section{Training and Implementation}
\subsection{Observation and Action Space}
The PPO policy is employed with an asymmetric input scheme, where the actor and the critic receive different observations. 

The actor observes a 70-dimensional vector $\mathbf{O}_\text{actor} \in \mathbb{R}^{70}$, 
consisting of joint positions $\mathbf{O}_\text{q\_pos} \in \mathbb{R}^{18}$, 
joint velocities $\mathbf{O}_\text{q\_vel} \in \mathbb{R}^{18}$, 
previous action $\mathbf{O}_\text{pre\_act} \in \mathbb{R}^{18}$, 
WBC command observation $\mathbf{O}_{\mathrm{WBC\text{-}cmd}} = \mathbf{c}_t^{\mathrm{WBC}} \in \mathbb{R}^{3+7}$,
base angular velocity $\mathbf{O}_\text{ang\_vel} \in \mathbb{R}^{3}$, 
and projected gravity vector $\mathbf{O}_\text{grav} \in \mathbb{R}^{3}$.

During training, the critic additionally receives privileged observations $\mathbf{O}_\text{priv}$, including base linear velocity $\mathbf{O}_\text{lin\_vel} \in \mathbb{R}^{3}$, linear and angular accelerations of the end effector ($\mathbf{O}_\text{EE\_lin\_acc}, \mathbf{O}_\text{EE\_ang\_acc} \in \mathbb{R}^{3}$), orientation deviation from gravity, Euclidean distance to the commanded end effector pose, and a local terrain height map $\mathbf{O}_\text{height} \in \mathbb{R}^{17 \times 11}$. These are used only during training to stabilize value estimation.

The action space is $\mathbf{a} \in \mathbb{R}^{18}$, corresponding to normalized target joint positions for 18 actuated joints (12 for the legs, 6 for the arm).

\vspace{-3mm}
\subsection{Objective Components} \label{reward}
The objective groups $R_{\mathrm{body}}$, $R_{\mathrm{loco}}$, and $R_{\mathrm{mani}}$ introduced in Section~\ref{prioritize} are constructed from the following task-specific components.

\textbf{Body objective terms:} $R_{\text{body}}$ includes regularization terms on body behaviors. Examples include undesired-contact avoidance, foot clearance, foot contact force regulation, body pose regulation, action smoothing, joint limit avoidance, and penalties on torque usage and joint accelerations.

\textbf{Locomotion objective terms:} $R_{\text{loco}}$ follows the standard Isaac Lab formulation~\cite{mittal2023orbit}, including exponential tracking reward of base linear and angular velocity.%

\textbf{Manipulation objective terms:} $R_{\mathrm{mani}}$ follows the decomposition introduced in Section~\ref{prioritize} and consists of coarse position tracking, fine position tracking, and orientation regulation terms. The position tracking terms are implemented as described in the previous section. The orientation objective includes a yaw-alignment term $r_{\mathrm{EE\text{-}yaw}}$ and an upright-orientation term $r_{\mathrm{EE\text{-}z\text{-}align}}$. The yaw-alignment term is defined as
\begin{equation}
r_{\mathrm{EE\text{-}yaw}}
=
\exp
\left(
-k e_{\mathrm{yaw}}^2
\right)
\end{equation}
\begin{equation}
\mathbf{d}_{\mathrm{EE}}^{W}
=~  
^{W}\boldsymbol{\mathcal{R}}_{\mathrm{EE}} \mathbf{e}_x
,
\qquad
\mathbf{d}_{\mathrm{cmd}}^{W}
=~
^{W}\boldsymbol{\mathcal{R}}_B~
^{B}\boldsymbol{\mathcal{R}}_{\mathrm{cmd}}
\mathbf{e}_x
\end{equation}
\begin{equation}
e_{\mathrm{yaw}}
=
\mathrm{atan2}
\left(
\hat{d}_{\mathrm{EE},x}^{W}
\hat{d}_{\mathrm{cmd},y}^{W}
-
\hat{d}_{\mathrm{EE},y}^{W}
\hat{d}_{\mathrm{cmd},x}^{W},~
\,~
(\mathbf{\hat{d}}_{\mathrm{EE}}^{W})^\top
\mathbf{\hat{d}}_{\mathrm{cmd}}^{W}
\right)
\end{equation}
where $\mathbf d_{\mathrm{EE}}^{W}$ and $\mathbf d_{\mathrm{cmd}}^{W}$ denote the end-effector and commanded forward directions in the world frame, respectively, $\mathbf{e}_x=[1,0,0]^\top$ is the local forward axis and $\hat{\mathbf d}$ denotes the horizontal projection normalized to unit length, with $\hat{d}_{(\cdot),x}$ and $\hat{d}_{(\cdot),y}$ its components. The yaw error is computed from the signed angle between their normalized horizontal projections. Here, $\boldsymbol{\mathcal{R}}$ denotes a rotation matrix.
The EE upright-orientation term is defined as
\begin{equation}
r_{\mathrm{EE-z-align}}
=
\exp(-k\bar{\theta}^2),
\qquad
\bar{\theta}
=
\arccos\left(\mathbf{z}^{\top}\mathbf{e}_z^W\right),
\end{equation}
where $\mathbf{z}$ denotes the end-effector local $z$-axis expressed in the world frame, and $\mathbf{e}_z^W=[0,0,1]^\top$ is the world-frame vertical direction. The term encourages gravity-aligned end-effector orientation.

\subsection{Training Protocol}

The policy is trained using PPO with the asymmetric actor--critic architecture in Isaac Lab. Training is performed in 4096 parallel environments on a NVIDIA 5090 GPU. Each episode runs for $1000$ policy steps at $50$~Hz ($20$~s; physics timestep 0.005 s, decimation = 4), without early termination. 

The base-velocity and end-effector-pose commands remain fixed during the first half of each episode and are jointly resampled once at the midpoint to represent a change in task intent. End-effector target sampling is conditioned on the base-velocity command. Under stationary commands, target positions are sampled from the full reachable workspace of the manipulator to encourage broad manipulation capability and diverse arm configurations. During locomotion, targets are sampled from a locomotion-oriented region within the reachable workspace, favoring targets that remain feasible under continuous body motion and terrain disturbances while retaining sufficient freedom for task-relevant end-effector regulation. This strategy concentrates training samples on locomotion-compatible targets while retaining task diversity.

\begin{figure*}
    \centering
    \includegraphics[width=0.95\linewidth]{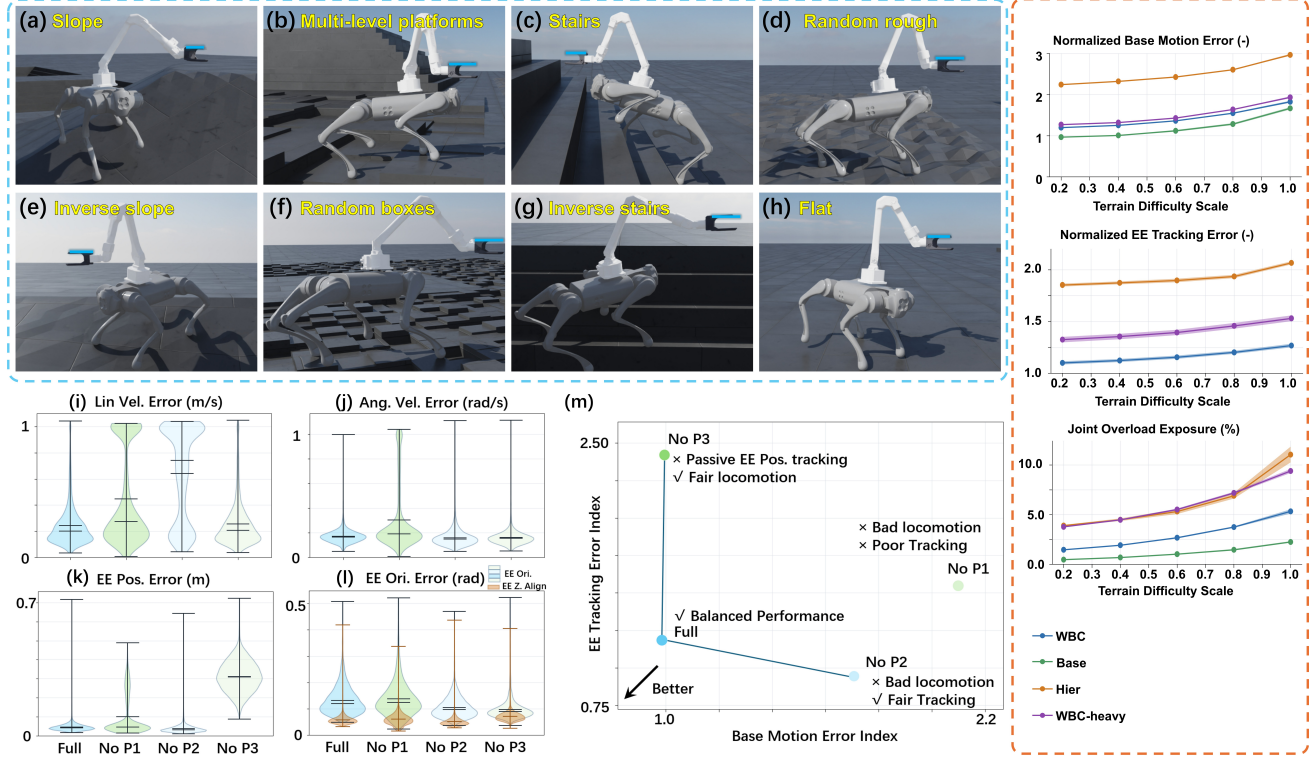}
    \caption{Ablation and performance evaluation of the proposed framework. 
    \textbf{Left:} (a)-(h) Representative behaviors of the full policy across terrains at difficulty 1.0. 
    (i)-(l) Distribution of locomotion and manipulation errors for different ablated variants (No P1-P3: corresponding Priority I-III regulation components disabled.).
    (m) locomotion-manipulation trade-off for the full policy and its ablations. 
    \textbf{Right:} Base Motion Error, EE Tracking Error, and joint overload exposure across terrain difficulty for WBC and the baselines. 
    Lower values indicate better performance.}
\vspace{-3mm}
    \label{fig:ablation}
\end{figure*}

Training uses procedurally generated terrains, including flat ground, slopes, stairs, boxes, and random obstacles. A terrain-level curriculum strategy gradually increases terrain difficulty from completely flat ground (difficulty 0.0) to the highest challenging terrain configuration (difficulty 1.0), as illustrated in Fig.~\ref{fig:ablation}(a--h).

Domain randomization is applied to dynamics parameters, payload conditions, and external disturbances to facilitate sim-to-real transfer. In particular, the PD stiffness and damping are scaled by $[0.8,1.3]$ and $[0.5,1.5]$, each arm-link mass is scaled by 1.0--1.8x, and an additional mass of $0$--$0.5$~kg is randomly attached to the end-effector to simulate varying payloads. 
An action delay of 0--3 control steps (0--60 ms) is also applied. External disturbances include an episode-long trunk wrench with each force and torque component sampled from $[-2,2]$ N and $[-2,2]$ N$\cdot$m, respectively, together with horizontal base-velocity perturbations in $[-0.5,0.5]$ m/s applied every 10--15s.

\section{Experiments and Results}

\subsection{Simulation Experiments}

\subsubsection{Ablation Study on Prioritized Whole-Body Objectives}\label{ablation}

To evaluate the proposed task-priority hierarchy, we conduct an ablation study on the prioritized whole-body objectives. The regulation components at different priority levels are selectively disabled to assess their contributions to whole-body performance.

All policies are trained under identical settings and evaluated on the most challenging terrain configuration using 1024 parallel simulation rollouts. The evaluation metrics include linear and angular base velocity tracking errors, end-effector position error, and end-effector orientation error.

For compact visualization, locomotion and manipulation performance are summarized using normalized composite metrics. For example, the Normalized EE Tracking Error is defined as
\begin{equation}
e_{\mathrm{EE}}
=
\sqrt{
\left(
\frac{e_{\mathrm{pos}}}
{e_{\mathrm{pos,ref}}}
\right)^2
+
\left(
\frac{e_{\mathrm{ori}}}
{e_{\mathrm{ori,ref}}}
\right)^2
},
\end{equation}
where $e_{\mathrm{pos}}$ and $e_{\mathrm{ori}}$ denote the end-effector position and orientation tracking errors, respectively, and the reference terms are normalization constants computed from the evaluation dataset.  The Base Motion Error is computed similarly using normalized linear and angular velocity tracking errors. These composite metrics are used only for compact visualization; the individual error terms are reported separately in Fig.~\ref{fig:ablation}(i--l).

Fig.~\ref{fig:ablation} (m) summarizes the overall trade-off between locomotion and manipulation using the composite Base Motion Error and EE Tracking Error metrics. The full policy achieves low errors in both metrics and lies closest to the lower-left corner, indicating the best overall balance among all variants.

Disabling Priority II component reduces the composite EE Tracking Error but substantially increases the composite Base Motion Error. This suggests that, without locomotion-priority objectives, the policy sacrifices mobility to maintain a stable end-effector pose. Although tracking performance remains favorable, the resulting policy lacks the locomotion capability required for complex terrain traversal.

In contrast, disabling Priority III component has little effect on locomotion performance but significantly increases end-effector tracking error. As shown in Fig.~\ref{fig:ablation} (k--l), the position error rises sharply while the orientation error remains relatively low. This indicates that the policy can satisfy orientation constraints while staying near locally comfortable arm configurations, but fails to actively approach the target. Without Priority III component, the manipulator does not fully exploit its workspace for goal-directed behavior, resulting in poor position tracking. This highlights the role of Priority III in promoting effective end-effector tracking.

Disabling Priority I component degrades both locomotion and manipulation performance, indicating that body regulation and contact-related objectives provide the foundation for higher-level behaviors. Without this layer, both mobility and end-effector performance deteriorate.

\subsubsection{Performance under Terrain and Manipulator Variations}

The effects of terrain difficulty and manipulator-related factors were evaluated in two tests, each using 1024 parallel environments per condition. 

\textbf{Terrain difficulty.}
Four configurations were evaluated across terrain difficulty levels ranging from 0.2 to 1.0: the proposed whole-body control policy with the Lite Arm (WBC), a locomotion-only baseline without manipulation (Base), a hierarchical framework combining locomotion control with IK-based arm tracking (Hier), and a whole-body policy trained with the manipulator mass scaled to 3 kg (WBC-heavy). Performance was assessed using Base Motion Error, EE Tracking Error, and joint overload exposure. The first two are the composite metrics defined in Section~\ref{ablation}, while joint overload exposure measures the percentage of near-limit actuator torque ($|\tau|\geq95\%\tau_{\max}$).
As shown in the right side of Fig.~\ref{fig:ablation}, all methods exhibit increasing errors and actuator load as terrain difficulty increases, indicating stronger terrain-induced disturbances and more challenging locomotion conditions. The degradation is more pronounced for the manipulation-enabled configurations, indicating the additional demands associated with the manipulator and end-effector tracking task. This effect is particularly evident in joint overload exposure. Compared with the locomotion-only baseline, the manipulation-enabled configurations experience more frequent near-limit actuator operation, and the gap widens on more challenging terrains.
Among the manipulation-enabled methods, WBC consistently achieves lower base-motion error, EE tracking error, and joint overload exposure than Hier, indicating that coordinated whole-body control reduces the effects of arm--body coupling and terrain-induced disturbances. WBC-heavy exhibits higher joint overload exposure than WBC, particularly on difficult terrains, suggesting that reducing manipulator mass can alleviate actuator loading under the evaluated conditions.

\textbf{Arm-motion speed and payload.}
Repeated lateral end-effector motions were evaluated at 0.4 and 0.8 m/s with payloads of 0.2 and 0.5 kg. For Hier, the failure rate rose from 8.50\% under the slow/light condition to 13.77\% under the fast/heavy condition, while the EE root-mean-square error (RMSE) over successful rollouts increased from 0.070 to 0.108 m. In comparison, WBC produced only two failures over all 4096 rollouts and maintained an EE RMSE of 0.057--0.069 m. These results indicate that WBC is less sensitive to arm-motion speed and payload.

\subsection{Real-World Experiments}

\subsubsection{Representative Terrain Traversal Demonstrations}

\begin{figure*}
    \centering
    \includegraphics[width=0.8\linewidth]{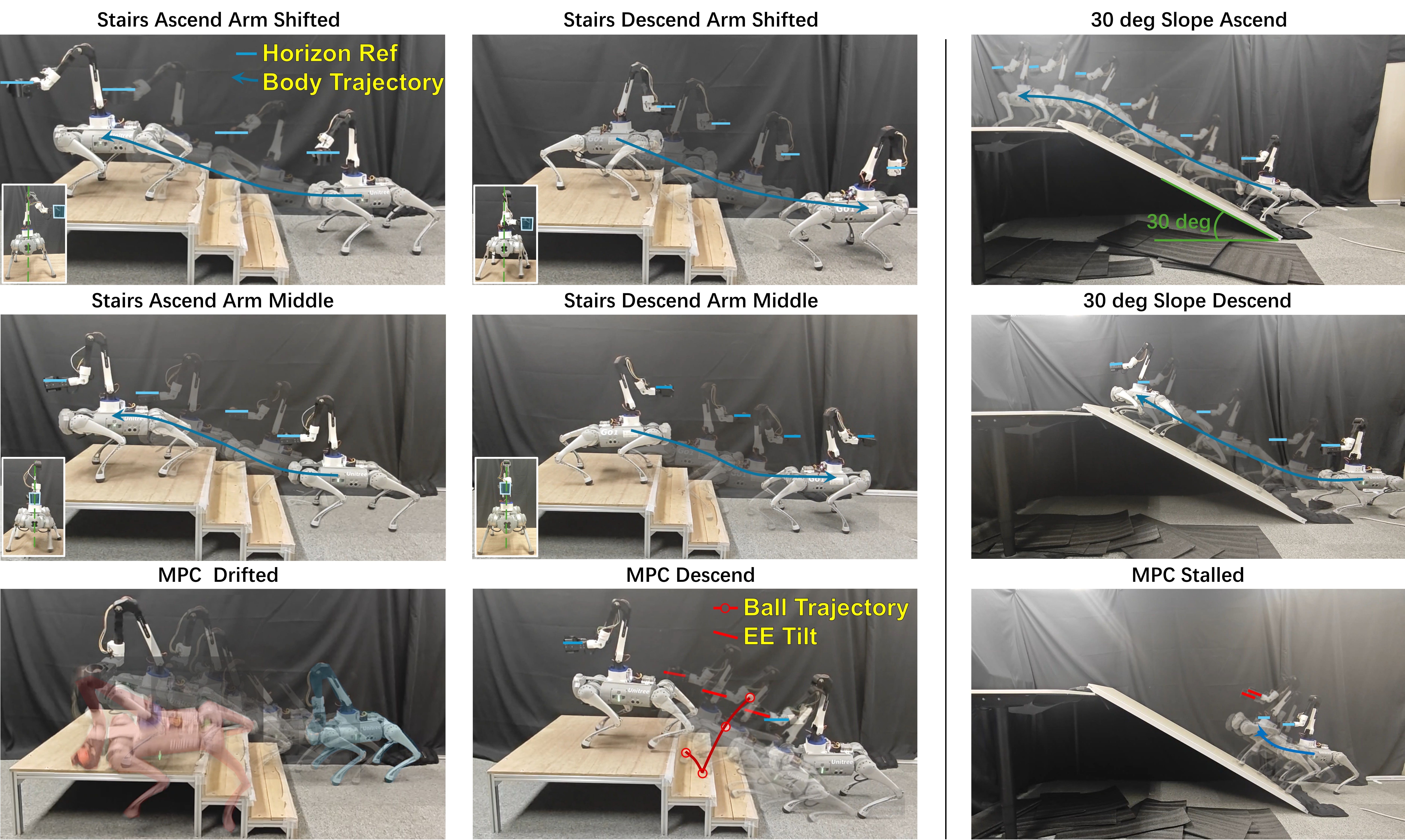}
    \caption{Representative rough-terrain traversal demonstrations. Blue lines indicate the horizontal direction of the end-effector. The proposed whole-body controller maintains a nearly level end-effector orientation during stair and slope traversal under different arm configurations, whereas the built-in MPC baseline exhibits larger orientation deviations and degraded terrain traversal performance.}
\vspace{-3mm}
    \label{fig:terrain-travelsal}
\end{figure*}
\begin{figure*}
    \centering
    \includegraphics[width=0.9\linewidth]{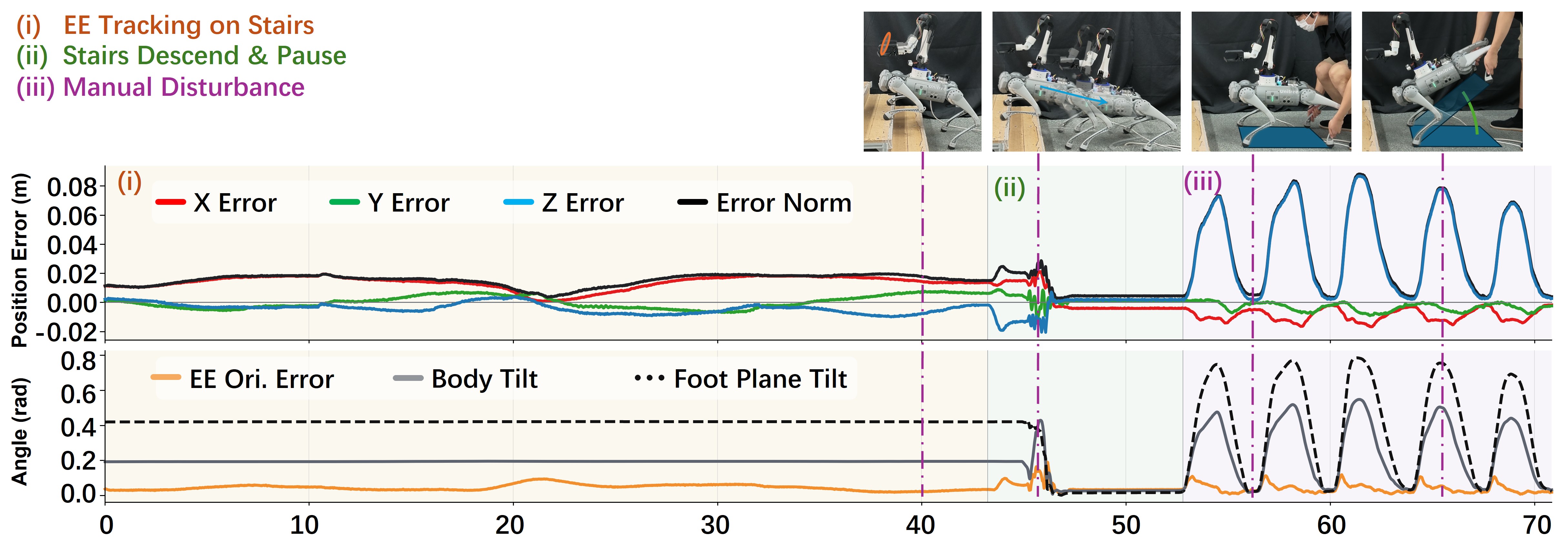}
    \caption{End-effector tracking on stairs and under manual disturbances. EE tracking errors are estimated via FK using measured joint states, with body and foot-plane tilts estimated from projected gravity and FK-based foot positions, respectively.}
    \vspace{-3mm}
    \label{fig:casestudy}
\end{figure*}
To provide representative real-world demonstrations of the proposed framework, the robot was tested on challenging uneven terrains, including a 30$^\circ$ slope and a sequence of 10 cm stairs, while carrying a container with two 36 mm elastic balls. The carried object serves as a visual indicator of end-effector disturbance during locomotion. The WBC policy runs onboard on an NVIDIA Jetson Xavier NX at 49.9 Hz, with a mean inference time of 1.53 ms and a sensor-to-command latency of 2.76 ms, while low-level joint control runs at 500 Hz.

As shown in Fig.~\ref{fig:terrain-travelsal}, the robot traverses both terrain types while maintaining coordinated locomotion and manipulation behaviors. During slope ascent, descent, and stair traversal, the end-effector remains approximately upright despite substantial body pitch and roll variations. The observed behavior is consistent with the proposed hybrid task-space formulation, where end-effector orientation is regulated with respect to gravity while position targets are represented in a body-centric coordinate system.

The stair demonstrations further illustrate the interaction between locomotion and manipulation objectives under varying arm configurations. Similar end-effector behavior is observed across different arm configurations, including forward-reaching and centered postures, despite the corresponding changes in robot mass distribution and body dynamics.

For comparison, the built-in MPC controller with a fixed arm posture exhibits larger body oscillations and noticeable lateral drift during terrain traversal. These body motions are directly reflected in the motion of the carried object, resulting in more pronounced end-effector disturbances.

Overall, these demonstrations qualitatively illustrate the behavior of the proposed whole-body controller during real-world terrain traversal. The observed results show coordinated locomotion and manipulation under complex terrain and varying arm configurations, providing representative examples of the proposed framework operating on challenging terrains.

\subsubsection{Quantitative Tracking and Whole-Body Compensation}

Fig.~\ref{fig:casestudy} quantitatively evaluates end-effector tracking and whole-body compensation during on-stair trajectory tracking and repeated manual pitching disturbances. The body-frame end-effector position is reconstructed by forward kinematics from the measured joint states, while its gravity-referenced orientation combines forward kinematics with the IMU-measured body attitude. Body and foot-plane tilts are obtained from projected gravity and a plane fitted to the forward-kinematics-based foot positions, respectively.
\begin{figure*}
    \centering
    \includegraphics[width=0.9\linewidth]{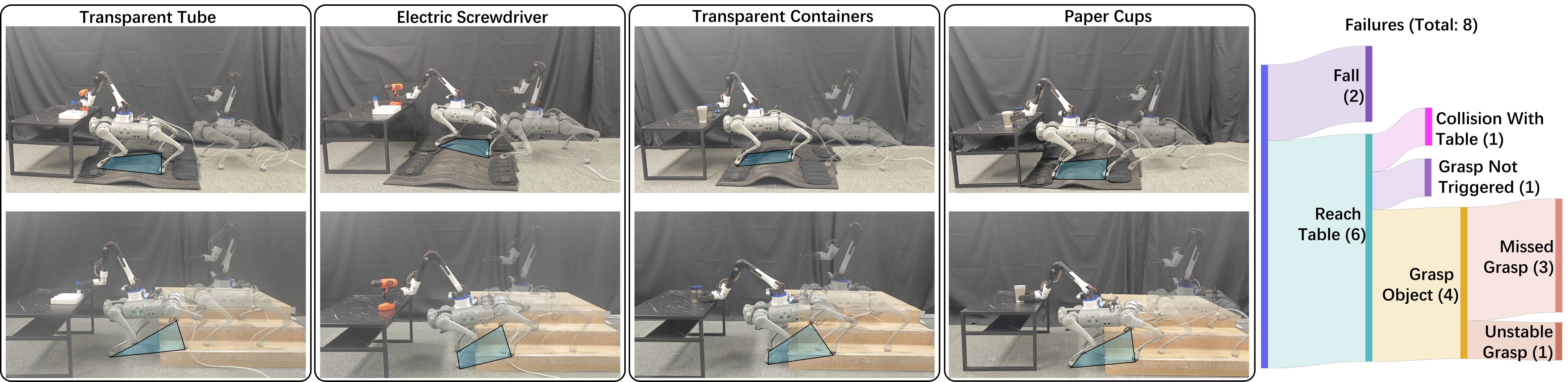}
    \caption{Autonomous rough-terrain grasping task demonstrations. 
    \textbf{Left:} execution sequences for the four target objects, each showing the robot traversing the terrain, approaching the table, and lifting the object; the top row is the deformable terrain, the bottom row the lateral stair descent, where the four feet occupy footholds at different heights. 
    \textbf{Right:} Sankey diagram of the eight failed trials, grouped by failure stage.
    }
    \label{fig:auto-grasp}
    \vspace{-3mm}
\end{figure*}
During on-stair trajectory tracking at a mean foot-plane tilt of 0.421 rad, the end-effector position and orientation errors were 0.016 $\pm$ 0.004 m and 0.045 $\pm$ 0.017 rad, respectively. Across the five manual disturbances, the average peak foot-plane tilt reached 0.752 $\pm$ 0.031 rad, while leg compensation limited the peak body tilt to 0.499 $\pm$ 0.037 rad. The arm further accommodated the residual pitching motion primarily through the z direction: the mean absolute x- and y-axis errors remained below 0.009 m, whereas the z-axis error was 0.034 $\pm$ 0.029 m. Meanwhile, the end-effector orientation error remained at 0.039 $\pm$ 0.024 rad. These results demonstrate coordinated compensation across the legs, body, and arm, preserving end-effector orientation under large pitching disturbances.

\subsubsection{Grasping on Deformable and Uneven Terrains}
\paragraph{Semantic-Guided Autonomous Grasping Pipeline}
To demonstrate the practical applicability of the proposed framework, we integrate the learned whole-body controller into a semantic-guided autonomous grasping pipeline running on an external workstation (NVIDIA RTX 5090). Given a natural language instruction, a Vision-Language Model (Qwen3-VL-32B)  is queried once at task onset with the current RGB observation as additional input. It returns a short target description containing scene-specific attributes, such as color and shape, that distinguish the target from other objects. This phrase serves as the text prompt for GroundingDINO~\cite{liu2024grounding},  which localizes the object at $10$~Hz; the corresponding onboard depth measurement yields its three-dimensional position, which is transformed into the robot body frame and used as the task objective.

A reachability-aware planner then continuously outputs a whole-body command $\mathbf{c}_t^{\mathrm{WBC}}$, consisting of a desired base velocity $\mathbf{v}$ and an end-effector target pose $\mathbf{x}$, corresponding to
$\mathbf{c}_t^{\mathrm{loco}}$ and $\mathbf{c}_t^{\mathrm{EE}}$, respectively:

\begin{equation}
\mathbf{c}^{\mathrm{WBC}}_t =
\begin{cases}
\left[\mathbf{0}, \mathbf{x}_\mathrm{grasp}\right],
& \mathbf{s}_{\mathrm{reach}}({}^{B}\mathbf{p}_{o}) = 1, \\[2mm]
\left[\mathbf{v}_\mathrm{approach}, \mathbf{x}_\mathrm{hold}\right],
&  \mathbf{s}_{\mathrm{reach}}({}^{B}\mathbf{p}_{o})=0,
\end{cases}
\end{equation}

where $\mathbf{s}_{\mathrm{reach}}(\cdot)$ is the reachability test. While the target is unreachable the robot approaches it under $\mathbf{v}_{\mathrm{approach}}$ while holding the end-effector. Once reachability is confirmed over consecutive observations, the planner switches to grasp execution. The command is interpolated to the $50$~Hz control rate of the WBC policy, which runs onboard and generates the coordinated leg and arm motions.

The WBC policy receives a base-velocity command and an end-effector command but no visual input. The perception pipeline affects the policy only through the body-frame target position. 
This target is sampled independently of any perception pipeline during policy training. Replacing the upstream perception module therefore does not by itself require retraining the WBC policy, provided that the replacement produces the same target representation within the policy's training range and at a comparable update rate. 
This modularity concerns the controller interface and does not imply equivalent end-to-end performance across different perception systems.
\begin{table*}[t]
\centering
\caption{Qualitative comparison with representative quadruped loco-manipulation systems.}
\label{tab:comparison}
\begin{tabular}{lcccc}
\toprule
Ref &
Terrain Demonstration &
Task-space Formulation &
System Scale (kg) &
Manipulation on Rough Terrain \\
\midrule

RAKOMO \cite{risiglione2025rakomo}
& Flat Ground
& World-oriented
& $\sim$90
& $\times$ \\

RoLoMa \cite{ferrolho2023roloma}
& Ramp
& World-oriented
& $>$50
& $\times$ \\

DeepWBC \cite{fu2023deep}
& Rough Terrain
& Body-oriented
& $\sim$14.8
& $\times$ \\

Visual WBC \cite{liu2024visual}
& Rough Terrain
& Body-oriented
& $\sim$60
& $\times$ \\

MLM \cite{liu2025mlm}
& Stairs / Obstacles
& World-oriented
& 19.9
& $\checkmark$ \\

UMI-on-Legs \cite{ha2024umi}
& Outdoor Terrain
& World-oriented
& $\sim$18.4
& $\triangle$ \\

RoboDuet \cite{pan2025roboduet}
& Gravel / Small Slope
& Spherical
& 15.35
& $\times$ \\

\textbf{Ours}
& \textbf{30$^\circ$ Slope / Stairs / Complex Terrain}
& \textbf{Hybrid (World + Body)}
& \textbf{13.5}
& \textbf{$\checkmark$} \\

\bottomrule
\end{tabular}

\vspace{1mm}
\footnotesize{
$\checkmark$: manipulation tasks reported during terrain traversal; $\triangle$: manipulation on mild outdoor scenario without explicit terrain-traversal evaluation. Bold indicates the proposed system.
}
\vspace{-5mm}
\end{table*}
\paragraph{Grasping Trials and Failure Analysis}

To evaluate the proposed framework in realistic task settings, we conducted autonomous semantic-guided grasping experiments on two challenging terrain categories: deformable terrain and lateral stair descent, as shown in Fig.~\ref{fig:auto-grasp}.
The target object is specified through a natural-language query (e.g., 'hand over paper cup' or 'give me the electric screwdriver'). Each trial starts several meters away from the target object, traverses the terrain, approaches the target, and autonomously executes the grasp once the object enters the reachable workspace. A trial is successful only if the robot both traverses the terrain and lifts the target object clear of the surface. Four everyday objects with diverse geometries and appearances are involved, including a paper cup, a transparent glass container, a transparent tube, and a 600 g electric screwdriver. Twenty trials were performed under each terrain condition, for a total of 40 trials.

The deformable terrain is built from sandbags and sponges, introducing compliant contacts and contact uncertainty that were not encountered during simulation training; $17/20$ trials succeeded. In the lateral stair descent, the four feet continuously occupy footholds at different heights, producing large pitch and roll variations of the support polygon; $15/20$ trials succeeded. The overall success rate is $80\%$ ($32/40$).

The eight failures (Fig.~\ref{fig:auto-grasp}, right) are distributed across distinct stages of the task rather than concentrated in a single subsystem. Two occurred during traversal, where the robot lost balance on the steepest footholds. The other six occurred after the terrain had been crossed: four from inaccurate target localization on the transparent tube and container, one where the base did not stop in time and contacted the table edge, and one from a grasp that closed but did not retain the object during the lift. Grouped by cause, three failures involve motion control (the two falls and the collision with the table), while the remaining five stem from localization or grasp execution.

\subsection{Positioning Among Loco-Manipulation Systems}

Table~\ref{tab:comparison} compares representative quadruped loco-manipulation systems in terms of terrain demonstrations, task-space formulation, system scale, and manipulation during terrain traversal.

Recent quadruped loco-manipulation systems have demonstrated increasingly capable whole-body control. However, most are evaluated on flat ground, mild outdoor environments, or manipulation-focused tasks~\cite{fu2023deep,liu2024visual,liu2025mlm,pan2025roboduet,ha2024umi}.
In contrast, the proposed system emphasizes task-consistent whole-body loco-manipulation during challenging terrain traversal. As summarized in Table~\ref{tab:comparison}, few existing quadruped manipulation systems report task execution on steep slopes or highly complex terrains. Our platform demonstrates loco-manipulation on slopes up to $30^\circ$ and complex terrain  while maintaining task-consistent end-effector behavior.

From the perspective of task-space representation, existing approaches typically formulate manipulation objectives either in body-centric coordinates~\cite{fu2023deep} or world-centric coordinates~\cite{liu2025mlm,ha2024umi}. The proposed hybrid task-space formulation instead combines body-frame position tracking with world-frame-aligned orientation regulation, keeping the end-effector z-axis parallel to the world z-axis, and provides a practical compromise between kinematic feasibility and task consistency under terrain-induced disturbances.

Compared with existing quadruped manipulation systems summarized in Table~\ref{tab:comparison}, the proposed system achieves terrain-adaptive loco-manipulation on a compact 13.5 kg platform while maintaining task-consistent end-effector behavior on challenging terrain.
\vspace{-3mm}
\section{Conclusion}

This paper presents a reinforcement-learning-based whole-body control framework for quadruped manipulators operating on challenging terrain. By combining a hybrid task-space representation with task-prioritized whole-body learning, the proposed framework enables coordinated locomotion and manipulation while preserving task-consistent end-effector behavior. Simulation and real-world experiments on slopes, stairs, and deformable terrain demonstrate that coordinated whole-body control enables terrain-adaptive loco-manipulation while maintaining task-consistent end-effector behavior and terrain traversal capability on small-scale quadruped manipulators. 
Several limitations remain. Because external motion-capture coverage was unavailable in the uneven-terrain test area, the quantitative real-robot evaluation was based on onboard inertial and joint measurements together with forward kinematics.
In addition, the current validation is limited to a single robot platform and a representative but limited set of terrain and manipulation conditions.

Future work will focus on large-scale loco-manipulation data collection, improving hardware design, and extending the framework to more complex manipulation and transportation tasks in dynamic environments.

\bibliographystyle{Bibliography/IEEEtranTIE}
\bibliography{Bibliography/IEEEabrv,Bibliography/ref}\ %

\vspace{-1.2cm}
\begin{IEEEbiography}[{\includegraphics[width=1in,height=1.25in,clip,keepaspectratio]{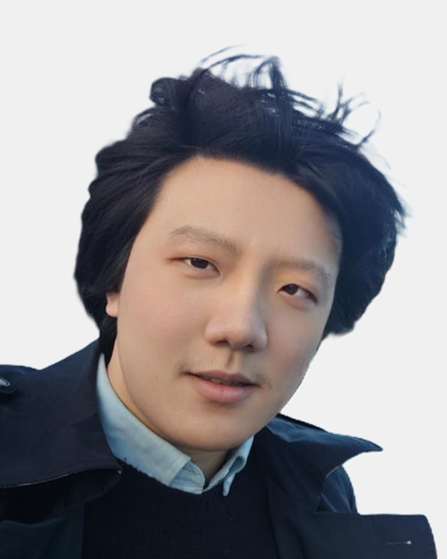}}]
{Yanzhou Jin}
received the B.Eng. degree from Beijing Institute of Technology, Beijing, China, in 2023, and the M.Sc. degree from Imperial College London, U.K., in 2024. He is currently pursuing a Ph.D. degree at the University of Tokyo, Japan. His research interests include reinforcement learning, whole-body control, and legged loco-manipulation.
\end{IEEEbiography}

\begin{IEEEbiography}[{\includegraphics[width=1in,height=1.25in,clip,keepaspectratio]{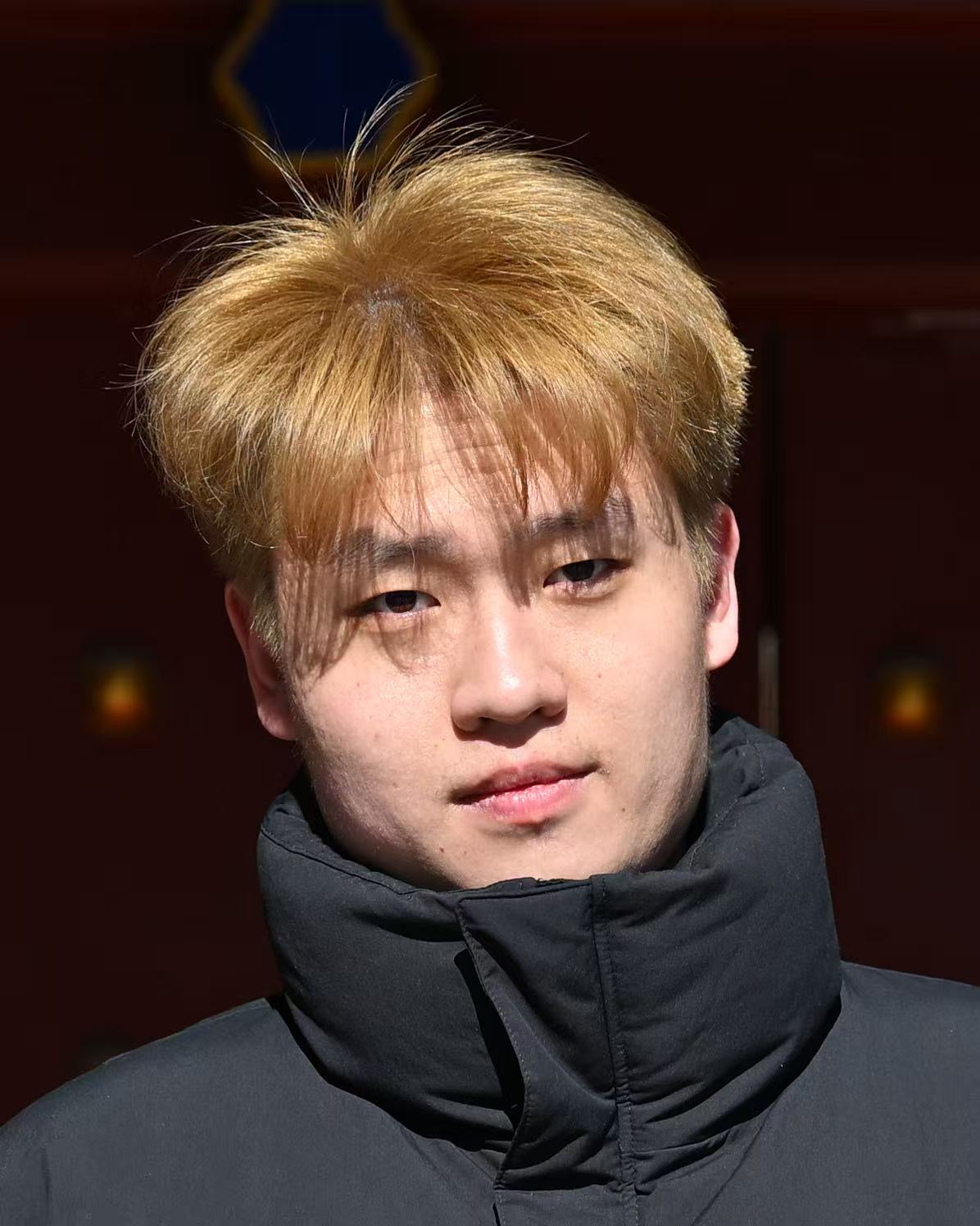}}]
{Fei Suo} received the B.E. degree in mechatronics engineering from Beijing Institute of Technology, Beijing, China, in 2023.
His research interests focus on soft robotics and, in particular, on their design, modeling, control, and perception.
\vspace{-1.2cm}
\end{IEEEbiography}

\begin{IEEEbiography}[{\includegraphics[width=1in,height=1.25in,clip,keepaspectratio]{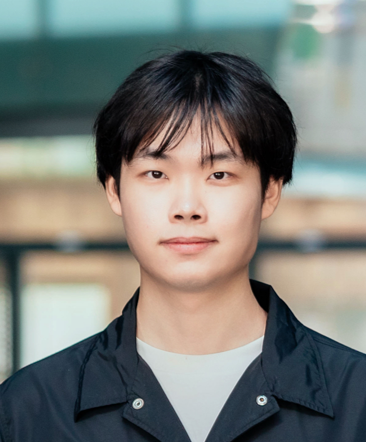}}]
{Chang Liu} received the M.S. degree from Kyoto University, and is currently pursuing the Ph.D. degree in The University of Tokyo, where his research interests include robot learning, reinforcement learning, and world models.
\vspace{-1.2cm}
\end{IEEEbiography}

\begin{IEEEbiography}[{\includegraphics[width=1in,height=1.25in,clip,keepaspectratio]{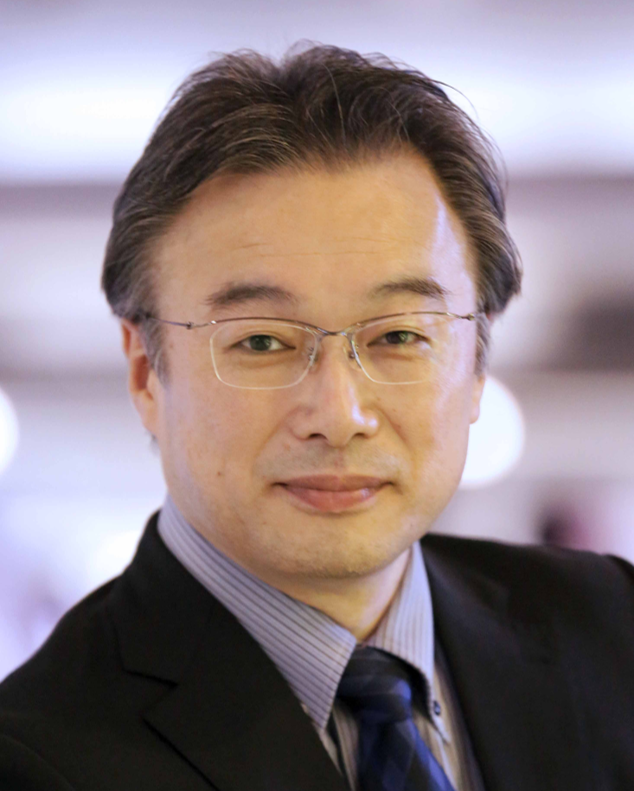}}]
{Yasuhisa Hasegawa} (Member, IEEE) received the Ph.D. degree from Nagoya University, Nagoya, Japan, in 2001. He is currently a Professor with Nagoya University, Nagoya, Japan. His research interests include physical human-robot interaction, assistive technology, and motion control.
\vspace{-1.2cm}
\end{IEEEbiography}

\begin{IEEEbiography}[{\includegraphics[width=1in,height=1.25in,clip,keepaspectratio]{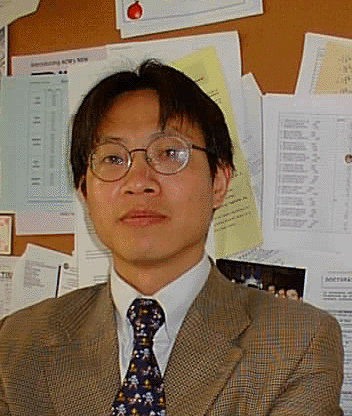}}]
{Liming Chen} (Senior Member, IEEE) received the M.S. and Ph.D. degrees in computer science from the University of Paris, Paris, France, in 1986 and 1989, respectively. He joined Ecole Centrale de Lyon, University of Lyon, Ecully, France, in 1998, where he is currently a Distinguished Professor. His research interests include machine learning, computer vision, pattern recognition, and multimedia analysis.
\vspace{-1.2cm}
\end{IEEEbiography}

\begin{IEEEbiography}[{\includegraphics[width=1in,height=1.25in,clip,keepaspectratio]{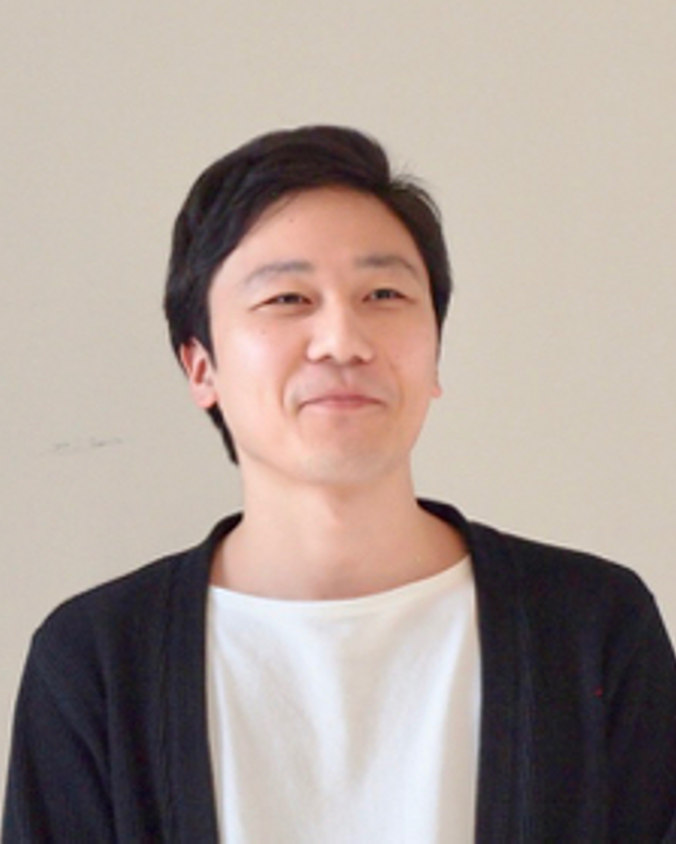}}]
{Yusuke Iwasawa} received the Ph.D. degree in engineering from The University of Tokyo, Tokyo, Japan, in 2017. He is currently an Associate Professor with the Graduate School of Engineering, The University of Tokyo. His research interests include machine learning, foundation models, and robot learning.
\vspace{-1.2cm}
\end{IEEEbiography}

\begin{IEEEbiography}[{\includegraphics[width=1in,height=1.25in,clip,keepaspectratio]{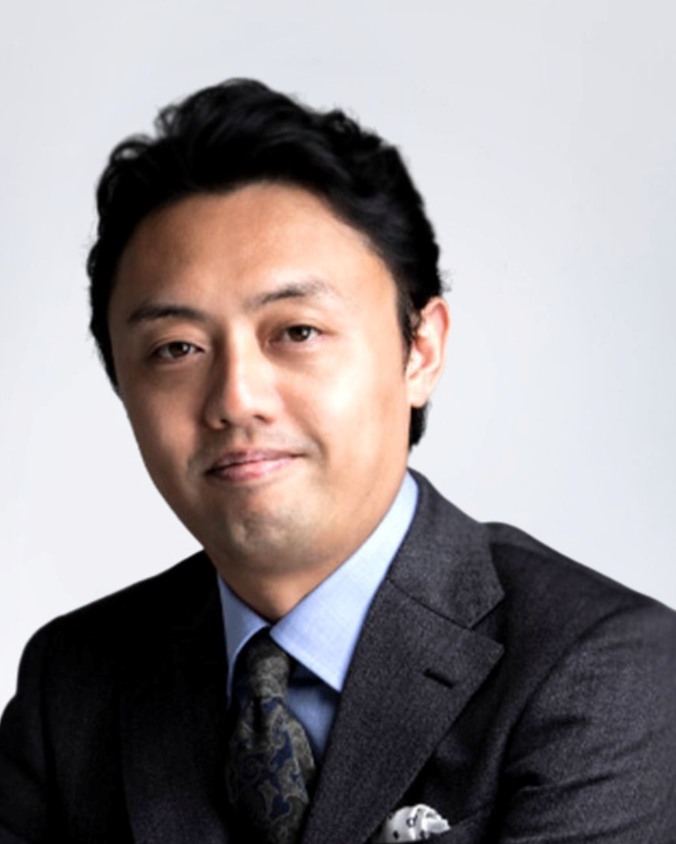}}]
{Yutaka Matsuo} received the Ph.D. degree in engineering from The University of Tokyo, Japan. He is currently a Professor with the Graduate School of Engineering, The University of Tokyo, where he leads the Matsuo-Iwasawa Laboratory. His research interests include deep learning, foundation models, and embodied AI. 
\vspace{-1.2cm}
\end{IEEEbiography}

\begin{IEEEbiography}[{\includegraphics[width=1in,height=1.25in,clip,keepaspectratio]{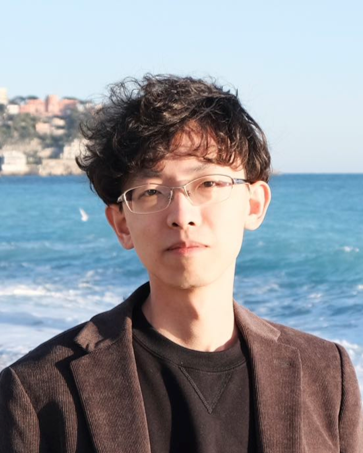}}]
{Yaonan Zhu} (Member, IEEE) received the B.S. degree from Northeastern University in 2015, the M.S. degree from the University of Science and Technology of China in 2018, and the Ph.D. degree in robotics from Nagoya University in 2021. He is currently a Project Lecturer at the University of Tokyo, Japan. His research interests include embodied AI, humanoid robotics, robot learning, and teleoperation. He received Best Paper Awards at IEEE ICMA 2024, MHS 2023, IEEE RO-MAN 2023, and the Best Paper Award Finalist at IEEE CBS 2023.
\end{IEEEbiography}

\end{document}